\documentclass{article}

\usepackage{PRIMEarxiv}
\usepackage{graphicx}
\usepackage[utf8]{inputenc} 
\usepackage[T1]{fontenc}    
\usepackage{hyperref}       
\usepackage{url}            
\usepackage{booktabs}       
\usepackage{amsfonts}       
\usepackage{nicefrac}       
\usepackage{microtype}      
\usepackage{lipsum}
\usepackage{fancyhdr}       
\usepackage{graphicx}       
\graphicspath{{media/}}     

\title{Twin identification over viewpoint change: A deep convolutional neural network surpasses humans
}

\author{
  Connor J. Parde \\
  School of Behavioral and Brain Sciences \\
  University of Texas at Dallas \\
  Richardson, TX\\
  \texttt{connor.parde@utdallas.edu} \\
   \And
  Virginia E. Strehle\\
  School of Behavioral and Brain Sciences \\
  University of Texas at Dallas \\
  Richardson, TX\\
   \And
  Vivekjyoti Banerjee \\
  University of Maryland Institute for Advanced Computer Studies \\
  University of Maryland \\
  College Park, MA\\
   \And
  Ying Hu\\
  School of Behavioral and Brain Sciences \\
  University of Texas at Dallas \\
  Richardson, TX\\
   \And
  Jacqueline G. Cavazos \\
  University of California Irvine\\
  School of Education \\
  Irvine, CA\\
   \And
  Carlos D. Castillo \\
  Whiting School of Engineering\\
  Johns Hopkins University \\
  Baltimore, MA\\
   \And
  Alice J. O'Toole\\
  School of Behavioral and Brain Sciences \\
  University of Texas at Dallas \\
  Richardson, TX\\
}

\begin{document}
\maketitle

\begin{abstract}

Deep convolutional neural networks (DCNNs) have achieved human-level accuracy in face identification (Phillips et al., 2018), though it is unclear how accurately they discriminate highly-similar faces. Here, humans and a DCNN performed a challenging face-identity matching task that included identical twins. Participants ($N=87$) viewed pairs of face images of three types: same-identity, general imposter pairs (different identities from similar demographic groups), and twin imposter pairs (identical twin siblings). The task was to determine whether the pairs showed the same person or  different people. Identity comparisons were tested in three viewpoint-disparity conditions: frontal to frontal, frontal to 45-degree profile, and frontal to 90-degree profile. Accuracy for discriminating matched-identity pairs from twin-imposters and general imposters was assessed in each viewpoint-disparity condition. Humans were more accurate for general-imposter pairs than twin-imposter pairs, and accuracy declined with increased viewpoint disparity between the images in a pair. A DCNN trained for face identification (Ranjan et al., 2018) was tested on the same image pairs presented to humans. Machine performance mirrored the pattern of human accuracy, but with performance at or above all humans in all but one condition. Human and machine similarity scores were compared across all image-pair types. This item-level analysis showed that human and machine similarity ratings correlated significantly in six of nine image-pair types [range $r=0.38$ to $r=0.63$], suggesting general accord between the perception of face similarity by humans and the DCNN. These findings also contribute to our understanding of DCNN performance for discriminating high-resemblance faces, demonstrate that the DCNN performs at a level at or above humans, and suggest a degree of parity between the features used by humans and the DCNN.

\end{abstract}


\section{Introduction}

Deep convolutional neural networks (DCNNs) now achieve human levels of accuracy on a variety of face recognition tests (for a review, see \cite{o2021face}). For high-quality frontal images of faces, DCNNs perform at the level of professional forensic face examiners, whose identification decisions can be used as evidence in court proceedings \cite{phillips2018face}. Human-machine comparisons have been common in face recognition research for more than a decade (cf., \cite{phillips2014comparison}). Most comparisons employ tests of face-identity matching (identity verification) whereby humans/machines decide whether two images show the ``same person'' or ``different people''. This task is commonly performed in security, law enforcement, and forensic applications \cite{frontex2012best, jain2002similarity, lynch2020face}. Because humans and machines are both good at face recognition, comparison studies require face stimuli that challenge both systems.
 
In the present study, we compare identity verification of identical twins by humans and a DCNN trained for face identification. Identical twins provide an extreme example of the difficult conditions humans (and machines) encounter in the real world when distinguishing between highly similar faces. Human expertise for faces has long been characterized as an ability to distinguish among large numbers of highly similar faces (e.g., \cite{carey1977piecemeal}). For humans, this expertise transcends the use of local feature information to rely also on subtle differences in the configural structure of the face (cf., \cite{maurer2002many}). Although this strategy is effective for most faces, it is less clear which features are most diagnostic for twin identification (cf. \cite{mousavi2021recognition,biswas2011study,srinivas2012analysis} and \cite{sundaresan2021monozygotic} for a review, see also Section \ref{diag_twins}).

Because identical twins comprise roughly 1 in 250 births, establishing the unique identity of individuals in identical twin pairs is also a pressing problem for face recognition systems used in security applications (e.g., passport control). This is especially true now that these systems have been scaled up to deal with millions of individual identities. Understanding the extent to which human- and/or machine-based face recognition is reliable for highly similar-faces, including twins, has both theoretical and applied value and can offer insight into the differences between the two face-identification systems.

\subsection{Distinguishing identical twins}\label{diag_twins}

Identical twins, referred to as monozygotic (MZ) twins, are a sibling pair that originate from the same egg and are fertilized by the same sperm. Therefore, the pair shares 100\% of their genetic make up \cite{gellman2013encyclopedia}, making it impossible to distinguish between a pair of identical twins based purely on DNA. Despite MZ twins sharing identical DNA, variations in appearance and/or susceptibility to diseases can emerge within the pair. These differences are due to epigenetics or to additional changes to the genetic sequence that affect how a gene is expressed \cite{weinhold2006epigenetics}. Epigenetic differences become more apparent as MZ twins age due to increased exposure to different environments. For example, 3-year-old MZ twins show a more similar pattern of DNA methylation, a common epigenetic process in which a methyl (CH\textsubscript{3}) group is either added or removed, compared to 50-year-old MZ twins \cite{fraga2005ballestar}. Finding variations in MZ twins does not always require extensive background knowledge of biology or genetics. There can also be phenotypic differences in the physical appearance of twins.

Techniques that detect variation in a twin pair that result as a byproduct of epigenetics or  physical differences between the two individuals have been used to differentiate identical twins. Iris texture, for example, does not depend on genetics. It has been found that iris texture is stable at as early as 8 months gestation, remains so across the lifespan \cite{sun2010study, li2014brief}, and can be differentiated by humans with more than 81\% accuracy with only 3 seconds of viewing time \cite{hollingsworth2010similarity}. Identical twins also have different fingerprints. By 7-months of age, the patterns on the fingers are fully developed and serve as a reliable measure of identity \cite{jain2002similarity,sun2010study}. Although iris scans and fingerprints can provide reliable biometrics for identical twins, neither is easily accessible for identity verification in security scenarios (e.g., passport control). 

Face recognition of MZ twin pairs relies on observable phenotypic variations within the twin pair. As noted, these phenotypic variations become more apparent as diverging epigenetic processes ensue. Changes in the appearance of the face can be due to natural effects of aging or by certain lifestyle-related behaviors. The face naturally changes across the lifespan \cite{farkas2013science}. For example, the skeletal structure of the face changes over time, altering an individual's face shape across the lifespan. Changes in the skin, like the deflation of subcutaneous fat and changes in musculature, also affect the appearance of the face. In addition, certain behaviors, such as smoking, can lead to more rapid changes in the appearance of the skin due to epidermal and dermal thinning, making the skin appear more droopy \cite{farkas2013science, guyuron2009factors,ricanek2013biometrically}. These features of facial aging combine to make twin faces more easily discriminable after the age of 40 \cite{phillips2011distinguishing,ricanek2013biometrically}.

Recently, computational approaches have been considered for extracting the face features that are diagnostic for discriminating identical twins. To locate ``critical features'' in pairs of images that depict the faces of identical twins, a Modified Scale Invariant Feature Transform \cite{lowe2004distinctive} was applied to determine mismatched facial key points between the image pairs \cite{mousavi2021recognition}. The points were then overlaid on five facial landmarks: eyes, eyebrows, nose, mouth, and face curve. The ``face curve'' (i.e., the outline of the lower face) contained the largest number of mis-matched points and was, therefore, considered the most diagnostic face region. Human ratings of the image pairs converged with the algorithm---face-curve ranked as the most diagnostic feature in approximately 35\% of twins’ face images \cite{mousavi2021recognition}. However, this finding is at odds with an earlier study in which facial markings (scars, moles, freckles) were rated as the most useful features (\cite{biswas2011study}, Section \ref{FR}). Moreover, using a twin discrimination algorithm based solely on facial markings showed that these markings are correlated across twin pairs \cite{srinivas2012analysis}, thereby making their value less clear.

\subsection{Computer-based Face Recognition of Monozygotic Twins}\label{FR}

In this section, we provide a brief history of studies examining the application of automatic face-recognition algorithms to the problem of identifying identical twins. For a comprehensive review of this literature, see \cite{sundaresan2021monozygotic}.

\subsubsection{Pre-DCNN Algorithms.}
Experiments testing the performance of commercial face-recognition algorithms between 2011-2014 concluded that face-recognition technology could not distinguish identical twins \cite{bowyer2016biometric}. Computer-based face-identification systems at that time typically used either principal component analysis (PCA) or hand-selected features to process face images \cite{turk1991eigenfaces, nechyba2007pittpatt}, and employed log-likelihood functions to reduce the error rate \cite{schneiderman2004learning}. 

The studies reviewed in this section rely almost exclusively on face images from the Notre Dame Twins dataset (ND-TWINS-2009-2010) \cite{pruitt2011facial}. This database contains images of identical twins, fraternal twins, and sibling pairs, taken at various poses and under different illumination conditions. Although another twin database was available at that time \cite{sun2010study}, ND-TWINS-2009-2010 includes racial diversity and contains far more identities. Notably, for some twin pairs, images are available from photographs taken in 2009 and 2010, thereby supporting time-lapsed recognition tests. The availability and quality of this dataset has spurred multiple studies of twin face recognition.

In one of the first papers to compare human and machine performance on twin identification \cite{biswas2011study}, human accuracy and strategy were studied with the goal of understanding (and potentially implementing in future work) successful methods used by human participants when distinguishing identical twins. Participants completed an identity-verification task in which they viewed pairs of identical-twin siblings (different-identity trials) and an equal number of same-identity image pairs. All pairs of images were taken under comparable illumination conditions. Participants rated the likelihood that the pair of images showed either the same person or identical-twin siblings, using a 5-point scale ranging from "1:Sure they are the same person" to "5: Sure they are identical twins". Humans performed substantially better when they were given more time to make a decision (average accuracy = 92.88\%) versus when they examined image pairs for only 2 seconds (average accuracy = 78.82\%). This time-dependency does not apply to non-twin faces, which are identified as accurately in 2 seconds as with unlimited time \cite{otoole2007}. Additionally, people were more accurate for twin pairs with facial markings. Among the computer algorithms tested, only one commercial algorithm (Cognitec) performed at a level that approached, but was below, human performance. 

In subsequent work with pre-DCNN systems, the focus of algorithm tests was on the extent to which image variation (illumination, pose, and expression) affected face identification in twins. Three commercially available face-verification systems (Cognitec 8.3.2.0, VeriLook 4.0, and PittPatt 4.2.1) and a face-verification system based on Local Region Principal Component Analysis (LRPCA) were tested on front-facing images of identical twins that varied in expression and illumination \cite{pruitt2011facial}. Of the commercial face-verification systems, two performed well on the controlled-illumination and neutral-expression test (Cognitec and VeriLook). This  was likely due to the sensitivity of these systems to high-resolution texture features in the face. PittPatt, which was optimized for small faces, performed poorly--as did the baseline LRPCA. However, the two face-identification systems that performed well for face images taken under the same conditions showed high
false-alarm rates when image conditions varied \cite{pruitt2011facial}. In related work \cite{phillips2011distinguishing}, the performance of 7 commercial algorithms was examined across variations in illumination, expression, gender, and age, for both the same-day and cross-year images \cite{paone2014double}. Three of algorithms tested were among the top submissions to the Multiple Biometric Evaluation (MBE) 2010 Still Face Track. Performance varied widely among the algorithms, though all algorithms performed less accurately when distinguishing identical twins versus non-twins. 

In general, the conclusion of this work is that twin recognition was achievable with pre-DCNN algorithms when the comparison images were well-matched. This is consistent with the more general face recognition literature at that time, which consistently showed for algorithms that the closer the  match in terms of image (e.g., similarity in pose, illumination, and expression) and appearance, the more accurate the performance of an algorithm. This rule also describes  human face recognition performance as a function of the similarity of image and appearance conditions in image pairs (e.g., \cite{braje1998illumination,o1998stimulus}). This early literature also showed that human participants identified identical twins far better than algorithms at that time.

\subsubsection{Deep-learning Approaches}
DCNNs \cite{krizhevsky2012imagenet}
have been remarkably successful in 
advancing the state of the art in automatic face recognition (e.g., \cite{chen2015end,dai2013convolutional,schroff2015facenet,ranjan2017all,taigman2014deepface,sun2014deep}). A strong advantage of these networks is their ability to generalize across image and appearance variation. There have been only a few  attempts to apply DCNN-related technologies to the problem of identifying twins \cite{mccauley2021identical,sun2018deep,ahmad2019deep, afaneh2017recognition}. These studies are difficult to compare to each other and to the previous literature, because they typically used twin datasets that differ from those used in the pre-DCNN era, and because these datasets are not entirely accessible. Moreover, they use DCNNs with diverse architectures and goals. In what follows, we briefly summarize these studies.

In one study \cite{afaneh2017recognition}, a combination of local feature extraction algorithm based on PCA, HOG, and LBP performed more accurately than an object-trained CNN on the ND-TWINS-2009-2010. In a subsequent study \cite{mccauley2021identical}, the goal was to create a baseline facial similarity measure between identical twins and to use this baseline to measure the impact of ``look-alike'' identities with no familial relationship. To that end, a larger face dataset was created by combining the twin dataset presented in \cite{paone2014double} and CelebA \cite{liu2018large}. A Siamese DCNN with a FaceNet architecture was trained to minimize the L2 distance between similar samples and to maximize L2 distance between dissimilar samples in the feature space. The output similarity score was the L2 distance between two samples in the feature space. The mean similarity score across all twin pairs served as the baseline facial similarity between identical twins. Next, the authors used the DCNN to process a large-scale non-twin dataset. They found that 6,153 out of the 15,455 identities (39.8\%) had one or more potential look-alikes, as defined by the aforementioned threshold. Furthermore, 288 identities (1.475\%) had one or more potential look-alike above the fourth quartile threshold. The authors suggest using this similarity score  to identify potential look-alikes in large-scale datasets to extract difficult cases for a facial recognition system or for intelligent morph pair generation.

In a second study \cite{sun2018deep}, a DCNN  was trained first on a large dataset, and was subsequently optimized to distinguish between identical twins. The model was tested on the Twin Days Festival Collection (TDFC), with high quality frontal face images of twins. The performance of the network was promising, with error rates of between 9.4 and 13.8\%. This study showed the plausibility of applying deep networks optimized for twin discrimination to the problem of identifying twins. One limit of the study, however, is that the dataset tested (TDFC) is not available to the public for replication and viewing. Thus, it is difficult to interpret the results of the experiments in terms of the challenge level of the stimuli.

\subsection{Experimental Goals}
The goal of the current project was to compare the identification accuracy of humans and a high-performing DCNN on a test that included identical twins, and to test performance across changes in viewpoint. As noted, a strong advantage of face identification DCNNs over previous generations of algorithms is that they show an ability to generalize identification over changes in viewpoint and illumination, (cf. for a review, \cite{o2021face}).  Accordingly, in Experiment 1, we designed a face-identity matching experiment in which humans judged the likelihood that pairs of frontal to frontal, frontal to three-quarter (45-degree) profile and frontal to full (90-degree) profile images showed the same person or different people. Image pairs showed either two different images of the same person, images of identical twins, or images of different people of the same gender, race, and approximate age. Face images were cropped to limit the available visual information to only the internal face. 

In Experiment 2, we tested a DCNN on the same task. We chose a network \cite{ranjan2017all} that performed at a level of accuracy comparable to professional forensic face examiners and super-recognizers \cite{phillips2018face}. We also tested whether there was a relationship between the perception of highly-similar images by humans and the DCNN by correlating human and machine ratings of similarity.

\section{Experiment 1 - Human Recognition of Identical Twins}
In Experiment 1, we measured twin identification performance in human participants, using the ND-TWINS-2009-2010 dataset. 

\subsection{Methods}
\subsubsection{Participants}
A total of 87 student participants were recruited from the University of Texas at Dallas. Participants were compensated with class credit in exchange for their time. For each viewpoint condition (frontal to frontal, frontal to 45-degree, frontal to 90-degree), there were 29 participants. Participants were required to be at least 18-years-old and have normal or corrected-to-normal vision. Eligibility was determined through self-report using a Qualtrics survey. All experimental procedures were approved by the UTD Institutional Review Board.\footnote{Note: the frontal condition was tested first, due to our initial assumption that the DCNN would not perform well enough to attempt an identification test on the 45- and 90-degree profile images. Thus, the 45- and 90-degree profile test for humans was carried out after the frontal test was complete, and so, human subjects were not assigned randomly to viewpoint groups.}

\subsubsection{Experimental Design}
Face-identity matching (identity verification) from pairs of images was tested as a function of the type of stimulus. Image pairs showed either the same identity (same-identity pairs) or different identities. In the latter case, the different identity pairs were either twin-imposter pairs or general-imposter pairs. {\it Same-identity pairs} consisted of two different images of the same identity. {\it Twin imposter pairs} consisted of identical twin siblings. {\it General imposter} pairs consisted of two images of different people who were not related to one another. Each of these pair types was tested in each of the three viewpoint conditions.

Identity-matching accuracy was measured by computing the area under the receiver operating characteristic curve (AUC) for two conditions: a.) same-identity pairs versus twin-imposter pairs and b.) same-identity pairs versus general imposter pairs.

\begin{figure}[t]
    \centering
    \includegraphics[width=\textwidth]{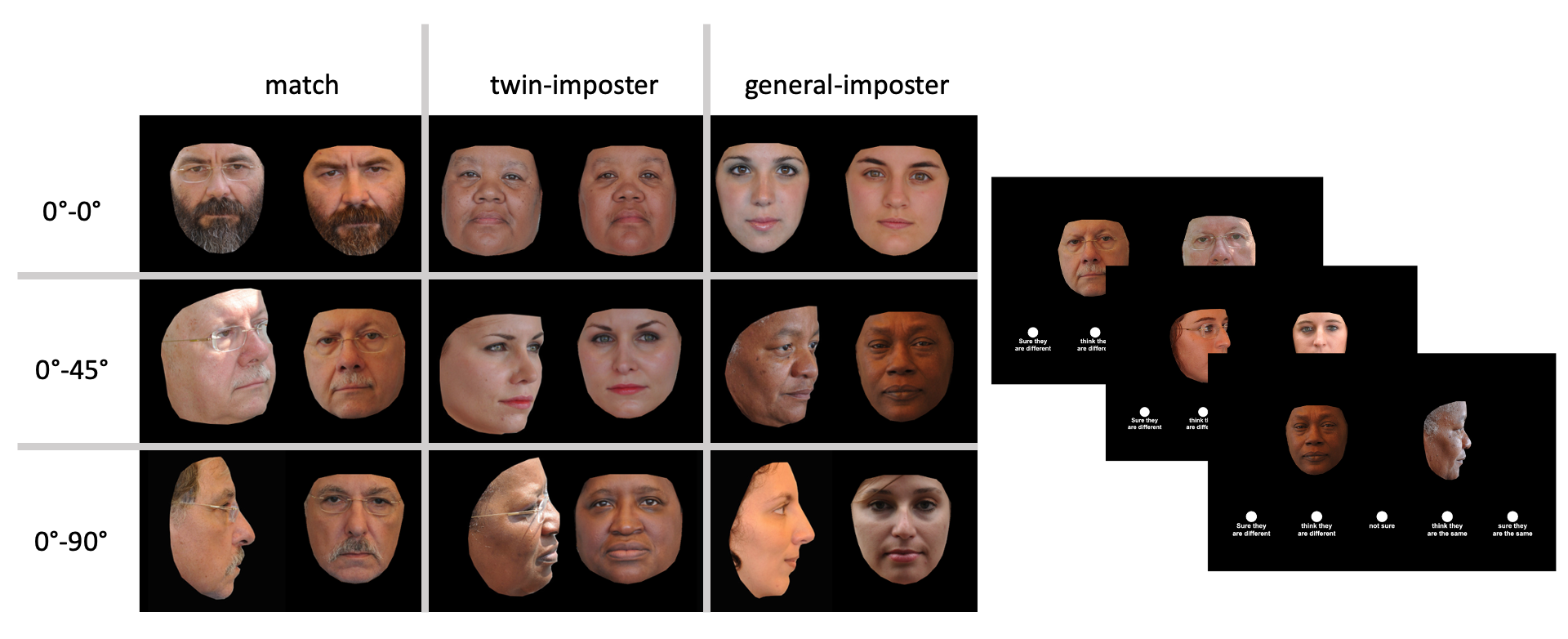}
    \caption{(Left) Examples of face-image pairs viewed during Experiment 1. The first column shows examples of same-identity pairs (match pairs), the second column shows examples of twin-imposter pairs (identical twins), and the third column shows examples of general-imposter pairs (unrelated individuals). The viewpoint-disparity conditions are shown along the rows, defined as frontal to frontal (top), frontal to 45-degree (middle), and frontal to 90-degree (bottom). (Right) The frame cascade depicts the experimental trials as seen by human participants. Each frame in the cascade depicts a different viewpoint-disparity condition.}
    \label{facePairsExampleImage}
\end{figure}

\subsubsection{Stimuli}
Images were selected from the ND-TWINS-2009-2010 database \cite{pruitt2011facial}. The  database contains 24,050 images of 435 different identities. Identities in the database are of sibling sets of one of the following types: identical twins, fraternal twins, non-twin siblings, and identical triplets. Identities in the database were photographed in both indoor and outdoor illumination conditions. Multiple viewpoints of each face were available for most identities, including:  -90-, -45-, 0-, 45-, and 90-degree views.

We included image pairs of Caucasian and African-American faces. Although the dataset included faces of other races, these groups did not include enough unique identities to be cross-balanced across experimental conditions. In each condition, we tested an equal number of matched-illumination pairs (indoor-indoor) and non-matched illumination pairs (indoor-outdoor). The resulting sample included 240 images of 200 identities, from which we formed 40 same-identity pairs, 40 twin-imposter pairs, and 40 general-imposter pairs, (120 trials). Across all trials, identities  appeared only once to prevent participants from becoming familiar with them. All faces were cropped to include only the inner-face region, with minimal hair visible.

Image pairs in the general imposter condition were comprised of two individuals of the same gender and race, with an age difference of no more than 8-years. To maximize the number of identity pairs available for inclusion, general imposter pairs were selected from the full range of identities in the data set regardless of an identity's sibling type.

For the frontal to 45-degree and frontal to 90-degree viewpoint disparity conditions, we began with the image pair used in the frontal to frontal condition. For each pair, we retained one (randomly selected) frontal image and replaced the second image with a 45- or 90-degree profile that matched the image parameters of the image being replaced. Example image pairs appear in Figure \ref{facePairsExampleImage}.

\subsubsection{Procedure}
Participants first completed a screening survey to determine eligibility. The survey confirmed that participants were at least 18 years old and had normal or corrected-to-normal vision. If participants satisfied the criteria, they were directed to an online informed consent form. Participants completed the informed consent form and then were given an access code to schedule a study time slot. During their scheduled time slot, the participant met with a research assistant over Microsoft Teams, using a link specific to the participant. 
 
The researcher briefly described the task by explaining to the participant that they would view a series of face pairs and and rate their certainty that the image pairs showed the same person or two different people\footnote{Note: The viewpoint conditions (frontal to frontal and frontal to profile) were tested at different times, there are some variations in the procedure. For the frontal to frontal condition, the researcher shared their computer screen with the participant over Microsoft Teams. Mouse control was transferred to the participant while they completed the experiment independently through the research assistant's computer. For the frontal to profile conditions, participants completed the same experiment format, but did so independently using Pavlovia.}. 

On each trial, a pair of face images appeared side-by-side on the screen. Participants were asked to determine whether the image pairs showed the same person or two different people using a 5-point scale. The response options included: ``1: Sure they are different, 2: Think they are different, 3:Not Sure, 4: Think they are the same, 5: Sure they are the same.'' Responses were reported by using a mouse to select the rating. The images and scale remained on the screen until the participant selected a response. The experiment was programmed in PsychoPy. The presentation order of the trials was randomized for each participant\footnote{Note: For the frontal to frontal condition, participants completed an exit survey that confirmed their demographics such as age and ethnicity. This survey was not employed for the frontal to profile conditions.}.

\subsection{Human Results}
\subsubsection{Accuracy}
Identity-matching accuracy was measured by computing the area under the receiver operating characteristic curve (AUC). For each participant in each viewpoint condition, an AUC was computed for: a) image pairs viewed in the general imposter condition, and b) image pairs viewed in the twin imposter condition. For both the general imposter condition and the twin imposter condition, correct identity verification responses were generated from same-identity image pairs. In the general imposter condition, false alarms were generated from image pairs that showed two distinct, unrelated identities. In the twin imposter condition, false alarms were generated from image pairs that showed  identical twins.  Because the distribution of correct verification responses was the same for both the general imposter condition and the twin imposter condition, the difference in the AUC between these conditions is due to identity-matching ability on the different-identity (twin-imposter and general-imposter) pairs.

\begin{figure} 
    \centering
    \includegraphics[width=\textwidth]{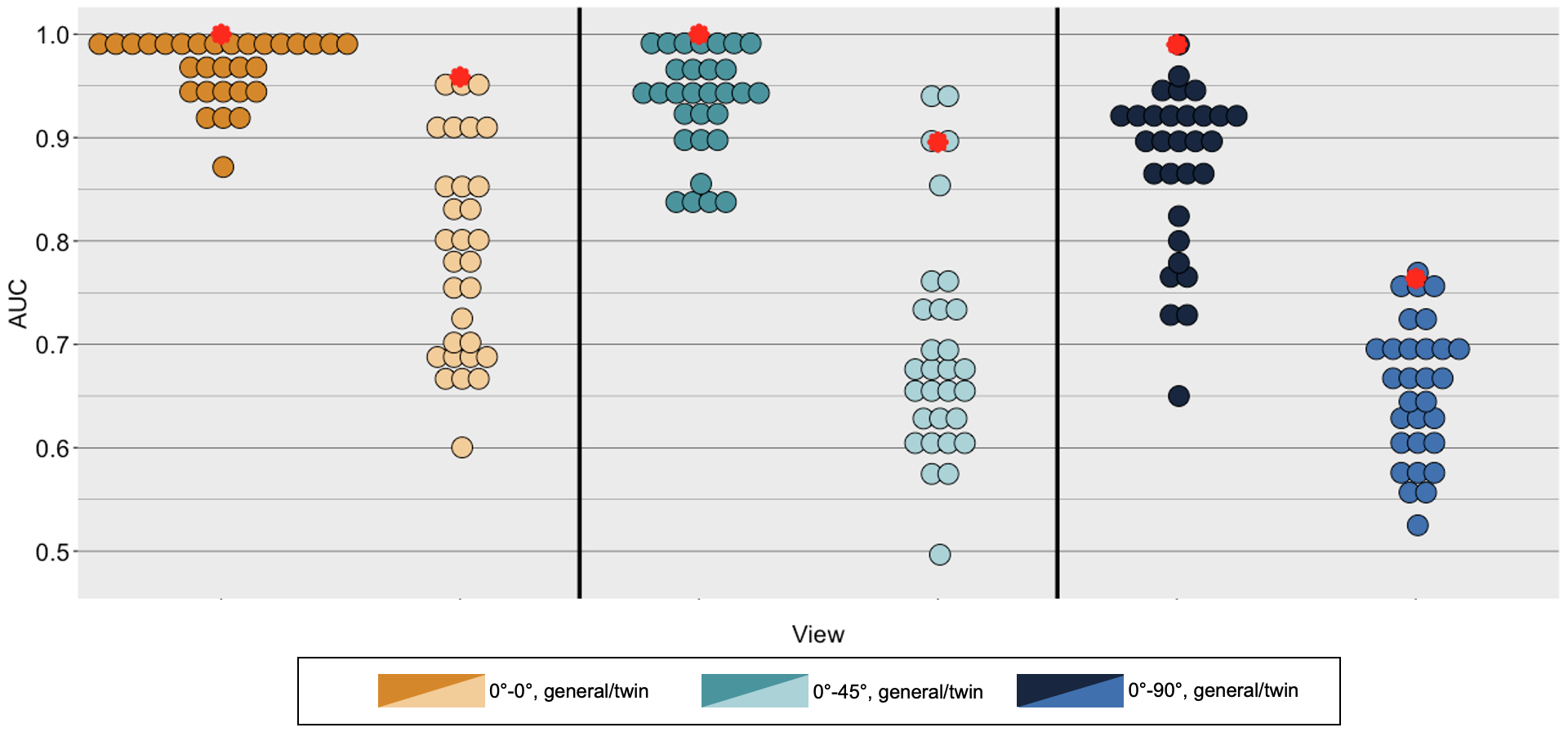}
    \caption{Identification accuracy for the DCNN and human participants. Both exhibited lower accuracy for twin identification, and both show a decline in performance as viewpoint disparity increases. The DCNN outperformed or performed at a level comparable to the highest performing humans in all conditions. (Red circles represent DCNN performance. Other circles represent individual human-participant accuracy.)  }
    \label{fig: Human AUC Score Distribution}
\end{figure}

Figure \ref{fig: Human AUC Score Distribution} shows a violin plot of human accuracy for the general and twin imposter conditions across viewpoint. The average AUC scores across humans by condition from left to right are as follows: general frontal to frontal (0.969), twin frontal to frontal (0.874), general frontal to 45-degree profile (0.933), twin frontal to 45-degree profile (0.691), general frontal to 90-degree profile (0.869), and twin frontal to 90-degree profile (0.652). More formally, the AUC data were submitted to a two-factor analysis of variance (ANOVA) with twin-condition (within-subjects) and viewpoint (between-subjects) as independent variables. As expected, performance was more accurate for the general imposter pairs than the twin imposter pairs, $F(1,84)= 649.84, p$ $\approx$ $0.00$, $\eta$ $^{2}_{G} = .28$. Performance also differed as a function of viewpoint disparity, $F(2,84)= 22.80, p$ $\approx$ $0.00$, $\eta$$^{2}_{G} = .67$. There was a significant interaction between viewpoint and twin condition,  $F(1,84)= 3.71, p$ $=$ $0.003$, $\eta$$^{2}_{G} = .02$). The pattern of means suggests that  performance declined more rapidly as a function of viewpoint change for the twin imposters than for the general imposters.

In summary, in all conditions, identity-matching was substantially more accurate for the general imposter condition than for the twin imposter condition. Accuracy decreased as a function of the viewpoint disparity between the images, with this decrease more pronounced for the twin imposter condition than the general imposter condition.

\section{DCNN Recognition of Identical Twins}

\subsection{Methods}
\subsubsection{Network Architecture}
For the algorithm test, we used a DCNN based on the ResNet-101 architecture \cite{ranjan2018crystal,he2016deep}. The network was trained on the Universe dataset, which is a web-scraped, ``in-the-wild'' dataset containing 5,714,444 images of 58,020 unique identities \cite{bansal2017s}. Images in this training dataset are sampled to include large variation in image attributes, including pose, illumination, resolution, and age \cite{bansal2017s}. The network contains 101 layers and employs skip-connections in order to maintain the amplitude of the error signal during training. Crystal loss is applied to ensure that $L_2$-norm is held constant during learning, and the alpha parameter is set to 50. When the fully trained network is complete, the output of the penultimate fully-connected layer is used to generate identity descriptors for each image processed through the network. The resulting network is a high-performing face-identification system that maintains accuracy across substantial changes in viewpoint, illumination, and expression (cf. \cite{Maze2018IARPAJB}).

\subsubsection{Similarity Scores}
To generate data that can be compared to humans, each face image presented to the human participants was processed through the DCNN to produce a descriptor for the image. To assess the similarity of the DCNN representation of the image pairs, we computed the cosine angle (i.e., normalized dot product) between the corresponding identity descriptors.

\subsection{Results: DCNN Identification accuracy}
DCNN-based similarity scores were generated for each of the image pairs viewed during the human data collection experiment. Identification accuracy for the DCNN was measured by computing the AUC for the similarity scores assigned to same-identity image pairs and different-identity images pairs. Correct responses were generated from image pairs showing the same identity, and false alarms were generated from image pairs showing different identities. DCNN performance is shown in Figure \ref{fig: Human AUC Score Distribution} as a red circle, overlaid on the individual human performance data.

For the general imposter condition, the DCNN obtained perfect identity-matching performance (AUC=1.0). For the identical-twin imposter condition, DCNN identity-matching performance remained high (AUC=0.96).

\section{Human and Machine Performance Comparison}

\begin{figure}
    \centering
    \includegraphics[width=\textwidth]{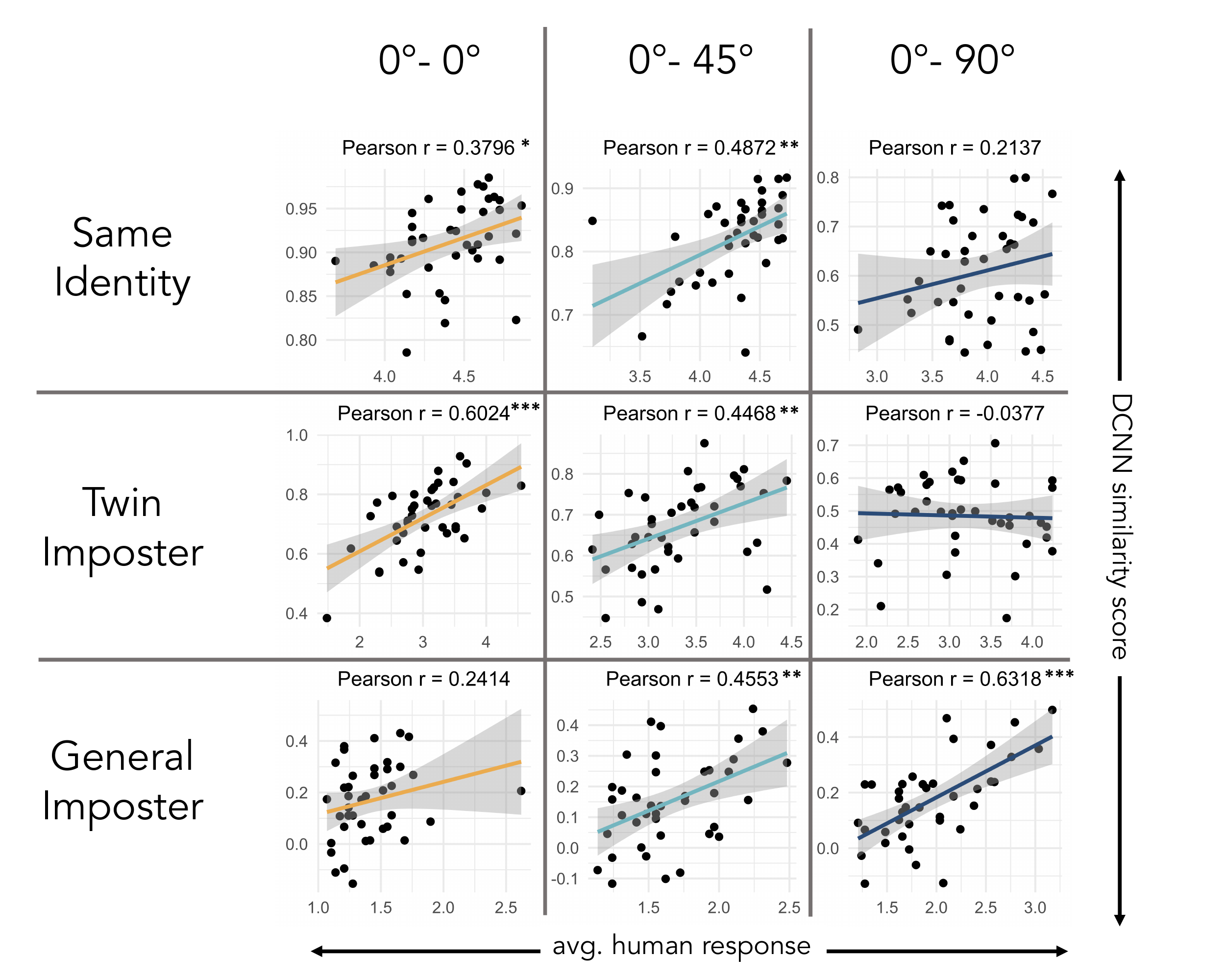}
    \caption{Plots depict correlations between human match ratings and the DCNN similarity score for each type of image pair. In 6 of 9 conditions, humans and DCNN ``similarity ratings'' correlated significantly.}
    \label{fig:correlations}
\end{figure}

To examine the relationship between human and DCNN ratings, Pearson product moment correlations were computed between average human ratings and DCNN similarity scores for all image pairs in the three viewpoint conditions. Figure \ref{fig:correlations} shows the scatter plots comparing average human response ratings on the $x$ axis and DCNN similarity scores on the $y$ axis with each point representing an image pair. Out of the 9 correlations, 6 were significant [range $r=0.38$ to $r=0.63$]. The majority significant correlations suggest that there is a relationship between the perceived similarity rating of the human and the similarity score generated by the DCNN. Combined, the results suggest a moderately strong relationship between the model' and humans' assessment of facial similarity.

\section{General Discussion}

Accurately distinguishing between identical twins requires the use of epigenetic biometric features that remain relatively stable across the lifespan. Although fingerprints and iris texture have been considered the most reliable biometrics for distinguishing between identical twins \cite{sun2014deep, sun2010study, hollingsworth2010similarity, jain2002similarity}, facial identification can also be used to distinguish identical twins, and has the advantage of being less invasive. Until the advent of DCNN algorithms, however, the best face identification algorithms performed well only when the images of twins were highly-controlled and matched for imaging conditions (e.g., both frontal pose, similar illumination conditions) \cite{biswas2011study, pruitt2011facial, paone2014double, phillips2011distinguishing}. Even in this controlled case, pre-DCNN algorithms perform poorly relative to humans on the task of discriminating identical twins \cite{biswas2011study}.
 
In the present study, we tested human and DCNN performance on a face-identification task that included identical twins, and which required identification over across large changes in viewpoint. Identical twins represent a special class of face stimuli that exhibit high similarity between different identities. The results of this study highlight the impressive progress of DCNNs over previous face-identification algorithms. We found that DCNN accuracy exceeded the accuracy of nearly all human participants tested in all conditions.  Although previous findings show that DCNNs perform at or above human levels of accuracy on a variety of face-identification tasks \cite{o2021face}, the present work extends the human-machine comparison results to the highly challenging task of twin identification over viewpoint change. 

One goal of this work was to gain insight into the relationship between face identification in humans and machines. The question of whether DCNNs perform in ways similar to humans can be considered at the level of the: a.) experimental conditions,  b.) individual participants, and c.) stimulus items. Starting with the experimental conditions, both the human and machine results showed the expected pattern of decreased accuracy for the twin-imposter pairs relative to the general-imposter pairs. Moreover, both also showed the expected accuracy decrease with increasing viewpoint disparity between the images in the comparison pair. 

From the perspective of the individual participants, a particularly notable aspect of human performance was the wide range of accuracy across individuals. In all conditions, some participants performed at the level of the machine, others moderately below that level, and still others {\it substantially} below the level of the machine. This is an important finding in the context of past work on face identification variability in law enforcement and security scenarios (cf., \cite{cavazos2020accuracy, phillips2018face,phillips2018face,white2015perceptual}). DCNN performance  was consistently at the top of the human performance distributions and is not subject to variability. Therefore, we might consider the ``machine'' to be similar to that a ``normal'' (untrained) person who is very good at the task of face recognition. However, the best human face recognizers are forensically trained face examiners and reviewers \cite{phillips2018face,white2015perceptual} and super-recognizers \cite{noyes2017super,phillips2018face}. These groups would likely be more accurate than the university students we tested. It would be of interest to test these special populations on the test administered here.

The strongest case for establishing a degree of parity between the face ``features'' used by humans and machines to discriminate identical twins can be made at the level of the individual stimulus items. We found significant item-based correlations between the average human response (1:sure same to 5:sure different) and the similarity of DCNN-generated image representations in 6 of the 9 experimental conditions. It is not possible to  determine why the human-machine judgments correlated more in some conditions than in others. Moreover, the conditions that showed little-to-no correlation do not easily lend themselves to speculation. We note only that two of the three conditions in which human and machine responses were unrelated involve the largest viewpoint disparity (match and twin imposter in the 0-90 viewpoint condition). That said, the highest human-machine correlation was obtained for the largest viewpoint disparity in the general imposter condition. It is possible that the difficulty of these conditions pushed participants into idiosyncratic strategies that may have added noise to the responses.  Notwithstanding, the significant correlations suggest that input from DCNN-based face identification systems can be used to predict the perceived similarity of an image pair as seen by human participants.

It remains a challenge going forward to come to a better understanding of the nature of the information captured in DCNN-generated representations of faces.  Some progress has been made on this question
from simulations that show that these representations retain detailed information both about the face identity and about the actual input image processed by the network \cite{parde2017face,hill2019deep}. Moreover, it is now known that the representation of image viewpoint is distributed across the 
units in the network output, whereas identity information is strongly represented both in individual units and across units \cite{parde2021closing}. A recent reviews provide an overview of what is known, and what still remains to be known \cite{o2021face}.

In summary, the present study tested human and machine performance on a face identification task involving identical twins, an extremely difficult task for humans and machines alike due to the high similarity of different-identity image pairs. The DCNN surpassed or performed at the level of the best humans across all experimental conditions, indicating that DCNNs now outperform the average human on a once human-dominated task. The results demonstrate that DCNNs are becoming highly accurate in more challenging face-identification contexts, more so than humans, suggesting that  difficult image-matching tasks (e.g., forensic applications) could benefit from joint human and DCNN decision making.

\section*{Acknowledgments}
Funding provided by National Eye Institute Grant R01EY029692-04 to AOT and CDC.

\bibliographystyle{unsrt}  
\bibliography{references}

\begin{thebibliography}{10}

\bibitem{o2021face}
Alice~J O'Toole and Carlos~D Castillo.
\newblock Face recognition by humans and machines: Three fundamental advances
  from deep learning.
\newblock {\em Annual Review of Vision Science}, 7:543--570, 2021.

\bibitem{phillips2018face}
P~Jonathon Phillips, Amy~N Yates, Ying Hu, Carina~A Hahn, Eilidh Noyes, Kelsey
  Jackson, Jacqueline~G Cavazos, G{\'e}raldine Jeckeln, Rajeev Ranjan, Swami
  Sankaranarayanan, et~al.
\newblock Face recognition accuracy of forensic examiners, superrecognizers,
  and face recognition algorithms.
\newblock {\em Proceedings of the National Academy of Sciences},
  115(24):6171--6176, 2018.

\bibitem{phillips2014comparison}
P~Jonathon Phillips and Alice~J O'toole.
\newblock Comparison of human and computer performance across face recognition
  experiments.
\newblock {\em Image and Vision Computing}, 32(1):74--85, 2014.

\bibitem{frontex2012best}
RDU Frontex.
\newblock Best practice operational guidelines for automated border control
  (abc) systems.
\newblock {\em European Agency for the Management of Operational Cooperation,
  Research and Development Unit,. https://bit. ly/2KYBXhz Accessed},
  9(05):2013, 2012.

\bibitem{jain2002similarity}
Anil~K Jain, Salil Prabhakar, and Sharath Pankanti.
\newblock On the similarity of identical twin fingerprints.
\newblock {\em Pattern Recognition}, 35(11):2653--2663, 2002.

\bibitem{lynch2020face}
Jennifer Lynch.
\newblock Face off: Law enforcement use of face recognition technology.
\newblock {\em Available at SSRN 3909038}, 2020.

\bibitem{carey1977piecemeal}
Susan Carey and Rhea Diamond.
\newblock From piecemeal to configurational representation of faces.
\newblock {\em Science}, 195(4275):312--314, 1977.

\bibitem{maurer2002many}
Daphne Maurer, Richard Le~Grand, and Catherine~J Mondloch.
\newblock The many faces of configural processing.
\newblock {\em Trends in cognitive sciences}, 6(6):255--260, 2002.

\bibitem{mousavi2021recognition}
Shokoufeh Mousavi, Mostafa Charmi, and Hossein Hassanpoor.
\newblock Recognition of identical twins based on the most distinctive region
  of the face: Human criteria and machine processing approaches.
\newblock {\em Multimedia Tools and Applications}, 80(10):15765--15802, 2021.

\bibitem{biswas2011study}
Soma Biswas, Kevin~W Bowyer, and Patrick~J Flynn.
\newblock A study of face recognition of identical twins by humans.
\newblock In {\em 2011 IEEE International Workshop on Information Forensics and
  Security}, pages 1--6. IEEE, 2011.

\bibitem{srinivas2012analysis}
Nisha Srinivas, Gaurav Aggarwal, Patrick~J Flynn, and Richard W~Vorder Bruegge.
\newblock Analysis of facial marks to distinguish between identical twins.
\newblock {\em IEEE Transactions on Information Forensics and Security},
  7(5):1536--1550, 2012.

\bibitem{sundaresan2021monozygotic}
Vinusha Sundaresan and S~Amala Shanthi.
\newblock Monozygotic twin face recognition: An in-depth analysis and plausible
  improvements.
\newblock {\em Image and Vision Computing}, 116:104331, 2021.

\bibitem{gellman2013encyclopedia}
Marc~D Gellman, J~Rick Turner, et~al.
\newblock {\em Encyclopedia of behavioral medicine}.
\newblock Springer New York, NY, USA:, 2013.

\bibitem{weinhold2006epigenetics}
Bob Weinhold.
\newblock Epigenetics: the science of change, 2006.

\bibitem{fraga2005ballestar}
Mario~F Fraga.
\newblock Ballestar e, paz mf, ropero s, setien f, ballestar ml, heine-suner d,
  cigudosa jc, urioste m, benitez j, boix-chornet m, sanchez-aguilera a, ling
  c, carlsson e, poulsen p, vaag a, stephan z, spector td, wu yz, plass c,
  esteller m.
\newblock {\em Epigenetic differences arise during the lifetime of monozygotic
  twins. Proc Natl Acad Sci USA}, 102:10604--10609, 2005.

\bibitem{sun2010study}
Zhenan Sun, Alessandra~A Paulino, Jianjiang Feng, Zhenhua Chai, Tieniu Tan, and
  Anil~K Jain.
\newblock A study of multibiometric traits of identical twins.
\newblock In {\em Biometric technology for human identification Vii}, volume
  7667, page 76670T. International Society for Optics and Photonics, 2010.

\bibitem{li2014brief}
Haiqing Li, Zhenan Sun, Man Zhang, Libin Wang, Lihu Xiao, and Tieniu Tan.
\newblock A brief survey on recent progress in iris recognition.
\newblock In {\em Chinese Conference on Biometric Recognition}, pages 288--300.
  Springer, 2014.

\bibitem{hollingsworth2010similarity}
Karen Hollingsworth, Kevin~W Bowyer, and Patrick~J Flynn.
\newblock Similarity of iris texture between identical twins.
\newblock In {\em 2010 IEEE Computer Society Conference on Computer Vision and
  Pattern Recognition-Workshops}, pages 22--29. IEEE, 2010.

\bibitem{farkas2013science}
Jordan~P Farkas, Joel~E Pessa, Bradley Hubbard, and Rod~J Rohrich.
\newblock The science and theory behind facial aging.
\newblock {\em Plastic and Reconstructive Surgery Global Open}, 1(1), 2013.

\bibitem{guyuron2009factors}
Bahman Guyuron, David~J Rowe, Adam~Bryce Weinfeld, Yashar Eshraghi, Amir Fathi,
  and Seree Iamphongsai.
\newblock Factors contributing to the facial aging of identical twins.
\newblock {\em Plastic and reconstructive surgery}, 123(4):1321--1331, 2009.

\bibitem{ricanek2013biometrically}
Karl Ricanek and Gayathri Mahalingam.
\newblock Biometrically, how identical are identical twins?
\newblock {\em Computer}, 46(3):94--96, 2013.

\bibitem{phillips2011distinguishing}
P~Jonathon Phillips, Patrick~J Flynn, Kevin~W Bowyer, Richard W~Vorder Bruegge,
  Patrick~J Grother, George~W Quinn, and Matthew Pruitt.
\newblock Distinguishing identical twins by face recognition.
\newblock In {\em 2011 IEEE International Conference on Automatic Face \&
  Gesture Recognition (FG)}, pages 185--192. IEEE, 2011.

\bibitem{lowe2004distinctive}
David~G Lowe.
\newblock Distinctive image features from scale-invariant keypoints.
\newblock {\em International journal of computer vision}, 60(2):91--110, 2004.

\bibitem{bowyer2016biometric}
Kevin~W Bowyer and Patrick~J Flynn.
\newblock Biometric identification of identical twins: A survey.
\newblock In {\em 2016 IEEE 8th international conference on biometrics theory,
  applications and systems (BTAS)}, pages 1--8. IEEE, 2016.

\bibitem{turk1991eigenfaces}
Matthew Turk and Alex Pentland.
\newblock Eigenfaces for recognition.
\newblock {\em Journal of cognitive neuroscience}, 3(1):71--86, 1991.

\bibitem{nechyba2007pittpatt}
Michael~C Nechyba, Louis Brandy, and Henry Schneiderman.
\newblock Pittpatt face detection and tracking for the clear 2007 evaluation.
\newblock In {\em Multimodal Technologies for Perception of Humans}, pages
  126--137. Springer, 2007.

\bibitem{schneiderman2004learning}
Henry Schneiderman.
\newblock Learning a restricted bayesian network for object detection.
\newblock In {\em Proceedings of the 2004 IEEE Computer Society Conference on
  Computer Vision and Pattern Recognition, 2004. CVPR 2004.}, volume~2, pages
  II--II. IEEE, 2004.

\bibitem{pruitt2011facial}
Matthew~T Pruitt, Jason~M Grant, Jeffrey~R Paone, Patrick~J Flynn, and Richard
  W~Vorder Bruegge.
\newblock Facial recognition of identical twins.
\newblock In {\em 2011 International Joint Conference on Biometrics (IJCB)},
  pages 1--8. IEEE, 2011.

\bibitem{otoole2007}
Alice~J. O'Toole, P.~Jonathon~Phillips, Fang Jiang, Janet Ayyad, Nils Penard,
  and Herve Abdi.
\newblock Face recognition algorithms surpass humans matching faces over
  changes in illumination.
\newblock {\em IEEE Transactions on Pattern Analysis and Machine Intelligence},
  29(9):1642--1646, 2007.

\bibitem{paone2014double}
Jeffrey~R Paone, Patrick~J Flynn, P~Jonathon Philips, Kevin~W Bowyer, Richard
  W~Vorder Bruegge, Patrick~J Grother, George~W Quinn, Matthew~T Pruitt, and
  Jason~M Grant.
\newblock Double trouble: Differentiating identical twins by face recognition.
\newblock {\em IEEE Transactions on Information forensics and Security},
  9(2):285--295, 2014.

\bibitem{braje1998illumination}
Wendy~L Braje, Daniel Kersten, Michael~J Tarr, and Nikolaus~F Troje.
\newblock Illumination effects in face recognition.
\newblock {\em Psychobiology}, 26(4):371--380, 1998.

\bibitem{o1998stimulus}
Alice~J O'toole, Shimon Edelman, and Heinrich~H B{\"u}lthoff.
\newblock Stimulus-specific effects in face recognition over changes in
  viewpoint.
\newblock {\em Vision research}, 38(15-16):2351--2363, 1998.

\bibitem{krizhevsky2012imagenet}
Alex Krizhevsky, Ilya Sutskever, and Geoffrey~E Hinton.
\newblock Imagenet classification with deep convolutional neural networks.
\newblock In {\em Advances in neural information processing systems}, pages
  1097--1105, 2012.

\bibitem{chen2015end}
Jun-Cheng Chen, Rajeev Ranjan, Amit Kumar, Ching-Hui Chen, Vishal~M Patel, and
  Rama Chellappa.
\newblock An end-to-end system for unconstrained face verification with deep
  convolutional neural networks.
\newblock In {\em Proceedings of the IEEE International Conference on Computer
  Vision Workshops}, pages 118--126, 2015.

\bibitem{dai2013convolutional}
Xiyang Dai.
\newblock A convolutional neural network approach for face identification.
\newblock In {\em Machine Learning, 30th International Conference on},
  volume~28, 2013.

\bibitem{schroff2015facenet}
Florian Schroff, Dmitry Kalenichenko, and James Philbin.
\newblock Facenet: A unified embedding for face recognition and clustering.
\newblock In {\em Proceedings of the IEEE conference on computer vision and
  pattern recognition}, pages 815--823, 2015.

\bibitem{ranjan2017all}
Rajeev Ranjan, Swami Sankaranarayanan, Carlos~D Castillo, and Rama Chellappa.
\newblock An all-in-one convolutional neural network for face analysis.
\newblock In {\em Automatic Face \& Gesture Recognition (FG 2017), 2017 12th
  IEEE International Conference on}, pages 17--24. IEEE, 2017.

\bibitem{taigman2014deepface}
Yaniv Taigman, Ming Yang, Marc'Aurelio Ranzato, and Lior Wolf.
\newblock Deepface: Closing the gap to human-level performance in face
  verification.
\newblock In {\em Proceedings of the IEEE conference on computer vision and
  pattern recognition}, pages 1701--1708, 2014.

\bibitem{sun2014deep}
Yi~Sun, Xiaogang Wang, and Xiaoou Tang.
\newblock Deep learning face representation from predicting 10,000 classes.
\newblock In {\em Proceedings of the IEEE conference on computer vision and
  pattern recognition}, pages 1891--1898, 2014.

\bibitem{mccauley2021identical}
John McCauley, Sobhan Soleymani, Brady Williams, John Dando, Nasser Nasrabadi,
  and Jeremy Dawson.
\newblock Identical twins as a facial similarity benchmark for human facial
  recognition.
\newblock In {\em 2021 International Conference of the Biometrics Special
  Interest Group (BIOSIG)}, pages 1--5. IEEE, 2021.

\bibitem{sun2018deep}
Xiaoxia Sun, Amirsina Torfi, Nasser Nasrabadi, M~Vatsa, R~Singh, and
  A~Majumdar.
\newblock Deep siamese convolutional neural networks for identical twins and
  look-alike identification.
\newblock In {\em Deep Learning in biometrics}, number~3, pages 65--83. CRC
  Press, 2018.

\bibitem{ahmad2019deep}
Belal Ahmad, Mohd Usama, Jiayi Lu, Wenjing Xiao, Jiafu Wan, and Jun Yang.
\newblock Deep convolutional neural network using triplet loss to distinguish
  the identical twins.
\newblock In {\em 2019 IEEE Globecom Workshops (GC Wkshps)}, pages 1--6. IEEE,
  2019.

\bibitem{afaneh2017recognition}
Ayman Afaneh, Fatemeh Noroozi, and {\"O}nsen Toygar.
\newblock Recognition of identical twins using fusion of various facial feature
  extractors.
\newblock {\em EURASIP Journal on Image and Video Processing}, 2017(1):1--14,
  2017.

\bibitem{liu2018large}
Ziwei Liu, Ping Luo, Xiaogang Wang, and Xiaoou Tang.
\newblock Large-scale celebfaces attributes (celeba) dataset.
\newblock {\em Retrieved August}, 15(2018):11, 2018.

\bibitem{ranjan2018crystal}
Rajeev Ranjan, Ankan Bansal, Hongyu Xu, Swami Sankaranarayanan, Jun-Cheng Chen,
  Carlos~D Castillo, and Rama Chellappa.
\newblock Crystal loss and quality pooling for unconstrained face verification
  and recognition.
\newblock {\em arXiv preprint arXiv:1804.01159}, 2018.

\bibitem{he2016deep}
Kaiming He, Xiangyu Zhang, Shaoqing Ren, and Jian Sun.
\newblock Deep residual learning for image recognition.
\newblock In {\em Proceedings of the IEEE conference on computer vision and
  pattern recognition}, pages 770--778, 2016.

\bibitem{bansal2017s}
Ankan Bansal, Carlos~D Castillo, Rajeev Ranjan, and Rama Chellappa.
\newblock The do's and don'ts for cnn-based face verification.
\newblock In {\em ICCV Workshops}, pages 2545--2554, 2017.

\bibitem{Maze2018IARPAJB}
Brianna Maze, Jocelyn~C. Adams, James~A. Duncan, Nathan~D. Kalka, Tim Miller,
  Charles Otto, Anil~K. Jain, W.~Tyler Niggel, Janet Anderson, Jordan Cheney,
  and Patrick Grother.
\newblock Iarpa janus benchmark - c: Face dataset and protocol.
\newblock {\em 2018 International Conference on Biometrics (ICB)}, pages
  158--165, 2018.

\bibitem{cavazos2020accuracy}
Jacqueline~G Cavazos, P~Jonathon Phillips, Carlos~D Castillo, and Alice~J
  O’Toole.
\newblock Accuracy comparison across face recognition algorithms: Where are we
  on measuring race bias?
\newblock {\em IEEE transactions on biometrics, behavior, and identity
  science}, 3(1):101--111, 2020.

\bibitem{white2015perceptual}
David White, P~Jonathon Phillips, Carina~A Hahn, Matthew Hill, and Alice~J
  O'Toole.
\newblock Perceptual expertise in forensic facial image comparison.
\newblock {\em Proceedings of the Royal Society B: Biological Sciences},
  282(1814):20151292, 2015.

\bibitem{noyes2017super}
Eilidh Noyes, P~Jonathon Phillips, and AJ~O’Toole.
\newblock What is a super-recogniser?
\newblock In {\em Face processing: Systems, disorders and cultural
  differences}, pages 173--201. Nova Science Publishers Inc, 2017.

\bibitem{parde2017face}
Connor~J Parde, Carlos Castillo, Matthew~Q Hill, Y~Ivette Colon, Swami
  Sankaranarayanan, Jun-Cheng Chen, and Alice~J O’Toole.
\newblock Face representations in deep convolutional neural networks.
\newblock In {\em IEEE Conference on Automatic Face and Gesture Recognition
  (FG), Gender and Discourse}, 2017.

\bibitem{hill2019deep}
Matthew~Q Hill, Connor~J Parde, Carlos~D Castillo, Y~Ivette Colon, Rajeev
  Ranjan, Jun-Cheng Chen, Volker Blanz, and Alice~J O’Toole.
\newblock Deep convolutional neural networks in the face of caricature.
\newblock {\em Nature Machine Intelligence}, 1(11):522--529, 2019.

\bibitem{parde2021closing}
Connor~J Parde, Y~Ivette Col{\'o}n, Matthew~Q Hill, Carlos~D Castillo,
  Prithviraj Dhar, and Alice~J O’Toole.
\newblock Closing the gap between single-unit and neural population codes:
  Insights from deep learning in face recognition.
\newblock {\em Journal of vision}, 21(8):15--15, 2021.

\end{thebibliography}

\end{document}